\documentclass[lettersize,journal]{IEEEtran}
\usepackage{amsmath,amsfonts}
\usepackage{algorithmic}
\usepackage{algorithm}
\usepackage{array}
\usepackage[caption=false,font=normalsize,labelfont=sf,textfont=sf]{subfig}
\usepackage{textcomp}
\usepackage{stfloats}
\usepackage{url}
\usepackage{verbatim}
\usepackage{graphicx}
\usepackage{cite}
\usepackage{multirow}
\usepackage{booktabs}
\usepackage{colortbl}
\usepackage{xcolor}
\usepackage{pifont}
\usepackage{float}
\usepackage{makecell}
\usepackage{adjustbox}
\usepackage{url}

\usepackage{algorithm}
\usepackage{algorithmic}
\usepackage{amsfonts}
\usepackage{amssymb}
\usepackage{marvosym}
\usepackage{multirow}
\usepackage{amsmath}
\usepackage{pifont}
\usepackage{colortbl}
\usepackage{adjustbox}
\usepackage{makecell}
\usepackage{booktabs}
\usepackage{float}
\usepackage{xcolor}
\usepackage{subfig}
\usepackage{tabularx}
\usepackage{marvosym}

\begin{document}

\title{InterMesh: Explicit Interaction-Aware \\End-to-End Multi-Person Human Mesh Recovery}

\author{Kaili Zheng*, Kaiwen Wang*, Xun Zhu, Chenyi Guo, Ji Wu, \IEEEmembership{Senior Member, IEEE}
\thanks{Manuscript received xxx, xxx; revised xxx, xxx. \textit{(Corresponding authors: Chenyi Guo, Ji Wu.)}}
\thanks{This work is supported by Beijing Natural Science Foundation (L242049) and Tsinghua University Initiative Scientific Research Program.}
\thanks{Kaili Zheng, Kaiwen Wang, Xun Zhu, Chenyi Guo and Ji Wu are with the Department of Electronic Engineering, Tsinghua University, Beijing 100084, China. Ji Wu  are also with the College of AI, Tsinghua University, Beijing 100084, China and Beijing National Research Center for Information Science and Technology. (e-mail: zkl25@mails.tsinghua.edu.cn; wkw23@mails.tsinghua.edu.cn; zhu-x24@mails.tsinghua.edu.cn; guochy@mail.tsinghua.edu.cn; wuji\_ee@mail.tsinghua.edu.cn)}
\thanks{Kaili Zheng and Kaiwen Wang contributed equally to this work.}}

\markboth{Journal of \LaTeX\ Class Files,~Vol.~14, No.~8, August~2021}%
{Shell \MakeLowercase{\textit{et al.}}: A Sample Article Using IEEEtran.cls for IEEE Journals}

\IEEEpubid{0000--0000/00\$00.00~\copyright~2021 IEEE}

\maketitle

\begin{abstract}
Humans constantly interact with their surroundings. Existing end-to-end multi-person human mesh recovery methods, typically based on the DETR framework, capture inter-human relationships through self-attention across all human queries. However, these approaches model interactions only implicitly and lack explicit reasoning about how humans interact with objects and with each other. In this paper, we propose InterMesh, a simple yet effective framework that explicitly incorporates human-environment interaction information into human mesh recovery pipeline. 
By leveraging a human-object interaction detector, InterMesh enriches query representations with structured interaction semantics, enabling more accurate pose and shape estimation. We design lightweight modules, Contextual Interaction Encoder and Interaction-Guided Refiner, to integrate these features into existing HMR architectures with minimal overhead. 
We validate our approach through extensive experiments on 3DPW, MuPoTS, CMU Panoptic, Hi4D, and CHI3D datasets, demonstrating remarkable improvements over state-of-the-art methods. Notably, InterMesh reduces MPJPE by 9.9\% on CMU Panoptic and 8.2\% on Hi4D, highlighting its effectiveness in scenarios with complex human-object and inter-human interactions. Code and models are released at \url{https://github.com/Kelly510/InterMesh}.
\end{abstract}


\section{Introduction}
\label{sec:introduction}

Human mesh recovery (HMR) aims to reconstruct 3D human meshes from RGB images, typically using a parametric model such as SMPL \cite{loper2015smpl}.
Due to its ability to capture detailed body shape and pose, HMR is widely used in applications such as animation production \cite{weng2019photo}, autonomous driving \cite{wang2023learning}, sports analysis \cite{liu2022posecoach}, and augmented or virtual reality \cite{xiu2022icon}.
Traditional approaches \cite{kolotouros2019learning,kocabas2020vibe,kocabas2021pare,xue20223d,li2022cliff,goel2023humans,cai2023smpler} often follow a two-stage pipeline: a person detector is first applied to localize individuals before each cropped region being fed into the HMR model. While this strategy enables focused feature extraction, it relies on only local information and fails to handle cases with severe occlusions or strong human-environment interactions.
Consequently, growing attention has shifted toward single-stage methods \cite{zhang2021body,sun2021monocular,sun2022putting,qiu2023psvt,sun2024aios,baradel2024multi,su2024sathmr} that directly estimate the meshes of all individuals in one pass by leveraging the full-scene context from the entire image.

\begin{figure*}
	\centering
	\includegraphics[width=\textwidth]{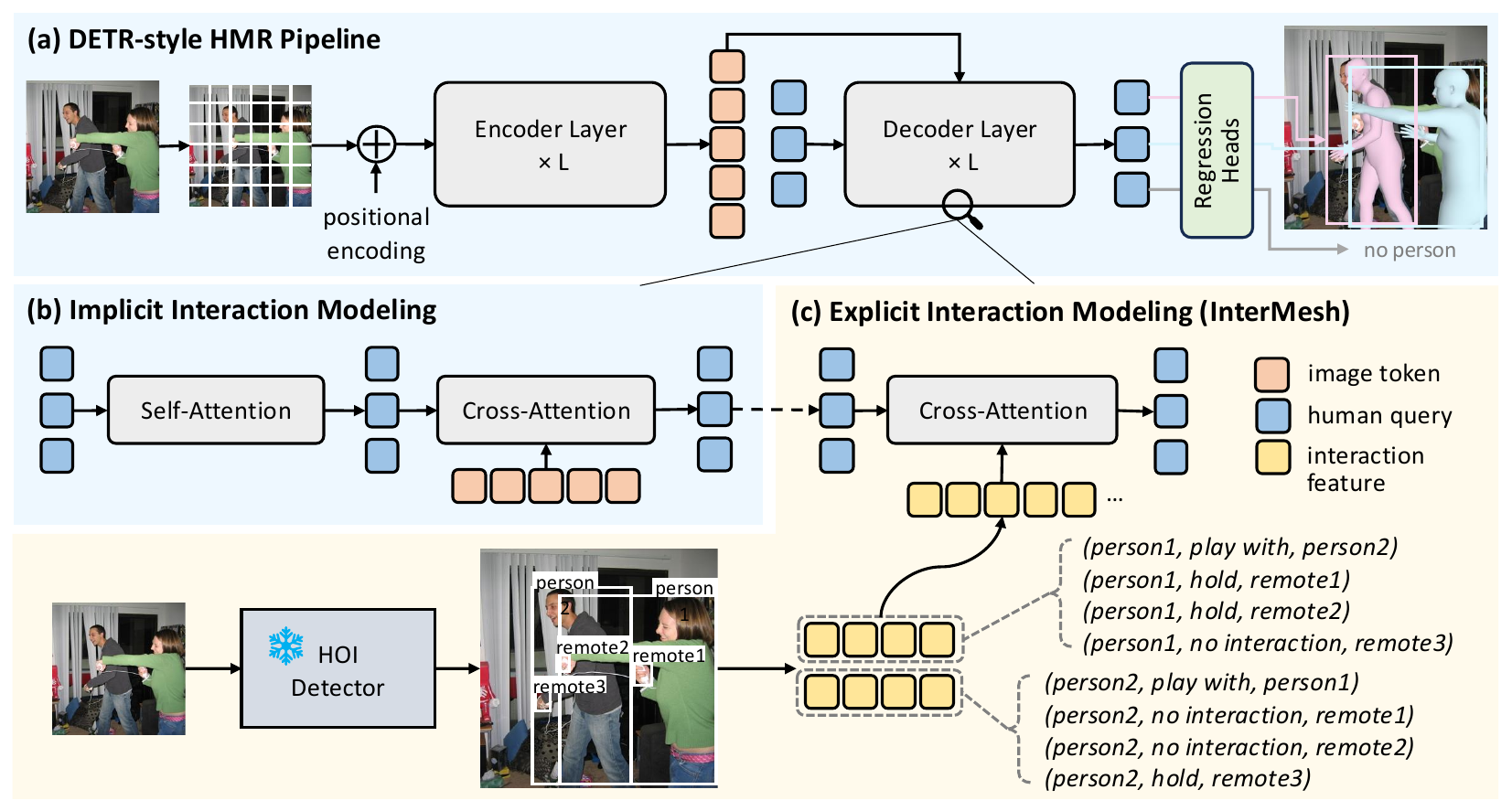}
	\caption{Implicit vs. explicit interaction modeling for DETR-style HMR. (a) DETR-style HMR employs a set of human queries (blue) to attend to image tokens extracted by the encoder (orange), where each query encodes the representation of an individual person and is subsequently used to regress the corresponding human parameters. (b) In prior DETR-style HMR models, each decoder layer consists of a self-attention module followed by a cross-attention module. Human queries first interact with one another through self-attention, and then attend to image tokens via cross-attention to extract person-specific features. Under this paradigm, interactions among different persons are modeled only implicitly through query self-attention. (c) InterMesh incorporates a pretrained Human-Object Interaction (HOI) detector to extract both human-object and inter-human interaction features (yellow). These interaction-aware features are injected into the decoder and used to refine human queries through cross-attention, enabling explicit modeling of human-environment interactions. By introducing rich semantic interaction cues, InterMesh provides stronger contextual guidance for downstream human mesh recovery.}
	\label{fig:teaser}
\end{figure*}

Recent state-of-the-art (SOTA) single-stage HMR methods \cite{qiu2023psvt,sun2024aios,su2024sathmr} usually follow a DETR-style \cite{carion2020end} pipeline shown in Fig.~\ref{fig:teaser}(a), where a set of learnable queries or queries instantiated from detected person-center tokens interact with image features to regress the parameters of the human body model in an end-to-end manner. Each query corresponds to one individual in the scene, enabling the model to detect and reconstruct multiple people simultaneously. In this pipeline, human queries alternate between self-attention among themselves and cross-attention with image features, shown in Fig.~\ref{fig:teaser}(b). The self-attention learns to model inter-human relationships implicitly from pose data. However, such operation is unable to explicitly identify the interaction semantics between persons. Such semantics, especially under conditions like heavy occlusion, can serve as strong priors for improving pose and shape estimation. Moreover, this pipeline focuses solely on inter-human relationships, overlooking richer human-environment interactions that are common in real-world scenarios. These limitations motivate the key question: \textit{can HMR models move beyond implicit reasoning and be explicitly guided to leverage interaction information — not only between people, but also between people and objects?}

A related line of work has explored interaction or scene context for human reconstruction. Interactive human reconstruction methods often leverage physical contact, proxemics, or joint human-object reconstruction to improve plausibility in close-contact scenarios~\cite{fieraru2023chi3d,huang2024closely,hassan2021populating,xie2022chore,tripathi2023deco,muller2024generative,fang2024capturing}. Scene-aware HMR methods further incorporate explicit 3D scene geometry, contact constraints, or egocentric scene context~\cite{shen2023learning,zhang2023probabilistic}. These methods highlight the importance of contextual reasoning, but they typically rely on contact priors, reconstructed or pre-scanned 3D scenes, object/scene meshes, or specialized egocentric settings. In contrast, our focus is on single-image, query-based multi-person HMR, where the challenge is how to introduce explicit interaction semantics into the end-to-end decoder without requiring 3D scene reconstruction.

Interestingly, the notion of modeling human interactions has been extensively studied in a different but related domain, Human-Object Interaction (HOI) detection. Unlike traditional HMR approaches that treat humans in isolation, HOI detection formulates interactions as semantic triplets involving a human, an object, and an action (e.g., \textit{person-holding-cup}), thereby emphasizing contextual understanding over individual-centric modeling. This perspective is highly complementary to HMR: rather than inferring pose and shape solely from appearance, models could benefit from the explicit interaction semantics provided by HOI reasoning, especially in cluttered or occluded scenes where pose cues are ambiguous. Despite the synergy between these two lines of work, existing HMR methods have largely overlooked the integration of HOI-based contextual priors.

To this end, we propose \textbf{InterMesh}, a framework that explicitly integrates human-environment \textbf{Inter}action features into end-to-end human \textbf{Mesh} recovery pipeline, as illustrated in Fig.~\ref{fig:teaser}(c). InterMesh leverages a pretrained zero-shot HOI detector \cite{lei2024ez} to extract both human-object and inter-human interaction cues, providing rich semantic context beyond visual appearance alone. Given a human query and its regressed bounding box, we first extract a set of associated interaction features using the HOI detector. These features are processed by a Contextual Interaction Encoder, which applies self-attention to model internal structure and semantic dependencies among interactions. The resulting contextualized features are then used by the Interaction-Guided Refiner, which integrates them into the human query via cross-attention. This allows the model to selectively attend to informative interaction cues, enhancing the quality of mesh estimation through structured environmental reasoning. By modeling interaction context explicitly, InterMesh bridges the gap between human-centric modeling and environmental semantics.

We evaluate InterMesh on the 3DPW, MuPoTS, CMU Panoptic, Hi4D, and CHI3D datasets. Quantitative results show significant improvements over prior SOTA methods across standard metrics. Qualitative comparisons further demonstrate that InterMesh leverages semantic cues from human-environment interactions to produce more accurate human mesh predictions, particularly in complex scenarios. These results suggest that, by bridging human mesh recovery with structured interaction understanding, InterMesh moves beyond traditional appearance-based reasoning and opens a new perspective on context-aware 3D human perception. It demonstrates that semantic cues, often overlooked in existing HMR pipelines, serve as powerful complementary signals for challenging scenarios involving occlusions, proximity, and collaborative actions. Additionally, ablation study is conducted to validate the contribution of each component within the InterMesh framework.

In summary, our contributions are as follows:
\begin{itemize}
	\item We propose InterMesh, the first method that explicitly incorporates human-environment interaction modeling into DETR-style HMR pipeline. By leveraging a pretrained zero-shot HOI detector, InterMesh introduces structured interaction tokens that encode both human-object and inter-human semantics, providing complementary contextual cues beyond appearance and geometry.
	\item We introduce Contextual Interaction Encoder and Interaction-Guided Refiner to integrate structured interaction features into human query refinement.
	\item Our method achieves SOTA performance on 3DPW, MuPoTS, CMU Panoptic, Hi4D, and CHI3D datasets, demonstrating that modeling humans within their surrounding context leads to more accurate and realistic mesh predictions. 
\end{itemize}

\section{Related Work}
\label{sec:related_work}

\subsection{Single-stage Human Mesh Recovery}
Differing from multi-stage approaches \cite{kolotouros2019learning,kocabas2020vibe,kocabas2021pare,xue20223d,li2022cliff,goel2023humans,cai2023smpler,patel2024camerahmr,hao2025twist} that first detect humans and then recover their meshes separately, single-stage methods concurrently recover the meshes of all people in the frame. 
% They take the whole image as input, allowing model to capture interactions between people and their surroundings. 
The high efficiency of single-stage models, which avoid multiple forward passes, and the potential of capturing more instance relationships has attracted increasing attention.
ROMP \cite{sun2021monocular} and BMP \cite{zhang2021body} predict human center maps and sample features around those centers to regress the human mesh. BEV \cite{sun2022putting} extends the center map to 3D space, capturing depth relationships between people. However, these center-map-based methods rely on simple processing of feature maps from CNN backbones \cite{he2016deep}, making it difficult to capture complex interactions.
Inspired by the success of DETR-style detectors \cite{carion2020end,zhu2020deformable,liu2022dab} in object detection, end-to-end HMR models following the DETR framework have shown competitive performance \cite{qiu2023psvt,sun2024aios,baradel2024multi,su2024sathmr}. After extracting low-level features with a backbone, PSVT \cite{qiu2023psvt} further leverages Vision Transformers to extract semantic features, and then uses these features together with human queries to decode each person's mesh. Multi-HMR \cite{baradel2024multi} initializes human queries from a predicted human center heatmap. AiOS \cite{sun2024aios} utilizes a multi-stage decoder to progressively refine the whole human mesh. These DETR-style methods use self-attention among human queries in the decoder to implicitly model inter-human interactions, but lack explicit semantic guidance for interaction modeling, leading to limited performance in scenarios with complex interaction. PromptHMR \cite{wang2025prompthmr} leverages multimodal prompts, such as interaction labels, to guide mesh estimation. However, it only uses binary contact indicators between individuals without capturing fine-grained interaction semantics. Moreover, due to its reliance on user-provided prompts and camera parameters, we do not include it in direct comparisons.

\subsection{Interactive Human Reconstruction}
Recent advances in human mesh recovery under interactive scenarios have been propelled by dedicated datasets such as CHI3D~\cite{fieraru2020chi3d,fieraru2023chi3d} and Hi4D~\cite{yin2023hi4d}. Existing methods mainly exploit interaction cues from physical contact, spatial proximity, or object-level reconstruction. Some methods explicitly model physical contact between humans~\cite{fieraru2023chi3d,huang2024closely}, while others extend to human-object interactions by jointly reconstructing human and object meshes to reason about their spatial relations~\cite{xie2022chore,tripathi2023deco}. BUDDI~\cite{muller2024generative} leverages a diffusion model on two-person proxemics, allowing reconstruction of two people in close interaction without explicit contact annotations, but remains limited to dyadic settings. Fang \textit{et al.}~\cite{fang2024capturing} proposes an interaction-aware self-attention mechanism to implicitly capture human-human interaction, showing promising results.
Concurrently, human-object interaction detection has seen significant progress, with transformer-based models and semantic reasoning methods effectively leveraging cues such as human intention, action categories and object affordances~\cite{xu2020interact,zhang2022efficient,lei2023efficient,fang2024hodn,mao2023clip4hoi,lei2024ez,chen2025ask}. These works suggest that semantic interaction cues can provide complementary information beyond appearance and geometry. However, existing interactive human reconstruction methods either focus on low-level physical contact or capture interactions implicitly through learned attention, and have not explicitly incorporated HOI-derived semantic interaction features into an end-to-end multi-person HMR decoder.

\subsection{Scene-aware Human Mesh Recovery}

Another related line of work incorporates 3D scene context into human mesh recovery. POSA~\cite{hassan2021populating} learns human-scene interaction priors by predicting body-scene contact and semantic affordances, enabling plausible human placement in 3D scenes. SA-HMR~\cite{shen2023learning} recovers human meshes in pre-scanned 3D scenes by leveraging scene geometry and contact cues, while EgoHMR~\cite{zhang2023probabilistic} addresses egocentric human mesh recovery with a scene-conditioned probabilistic formulation to handle severe truncation and ambiguity. In addition, video-based global HMR methods combine human reconstruction with camera motion estimation or SLAM-style scene reasoning to recover humans in a shared world coordinate frame~\cite{yuan2022glamr,ye2023decoupling,shin2024wham,zhao2024synergistic}.
These scene-aware methods improve geometric and physical plausibility by exploiting explicit 3D scene structure, but they typically rely on pre-scanned scenes, reconstructed scene geometry, egocentric video, camera trajectories, or optimization over temporal observations. In contrast, InterMesh targets single-image, query-based multi-person HMR from RGB input. Rather than reconstructing or requiring a 3D scene, our method introduces explicit human-environment interaction tokens from an HOI detector, covering both human-human and human-object relations, and injects these semantic cues into human query refinement.

\section{Preliminaries}
\label{sec:preliminaries}

\begin{figure*}[htbp]
	\centering
	\includegraphics[width=\textwidth]{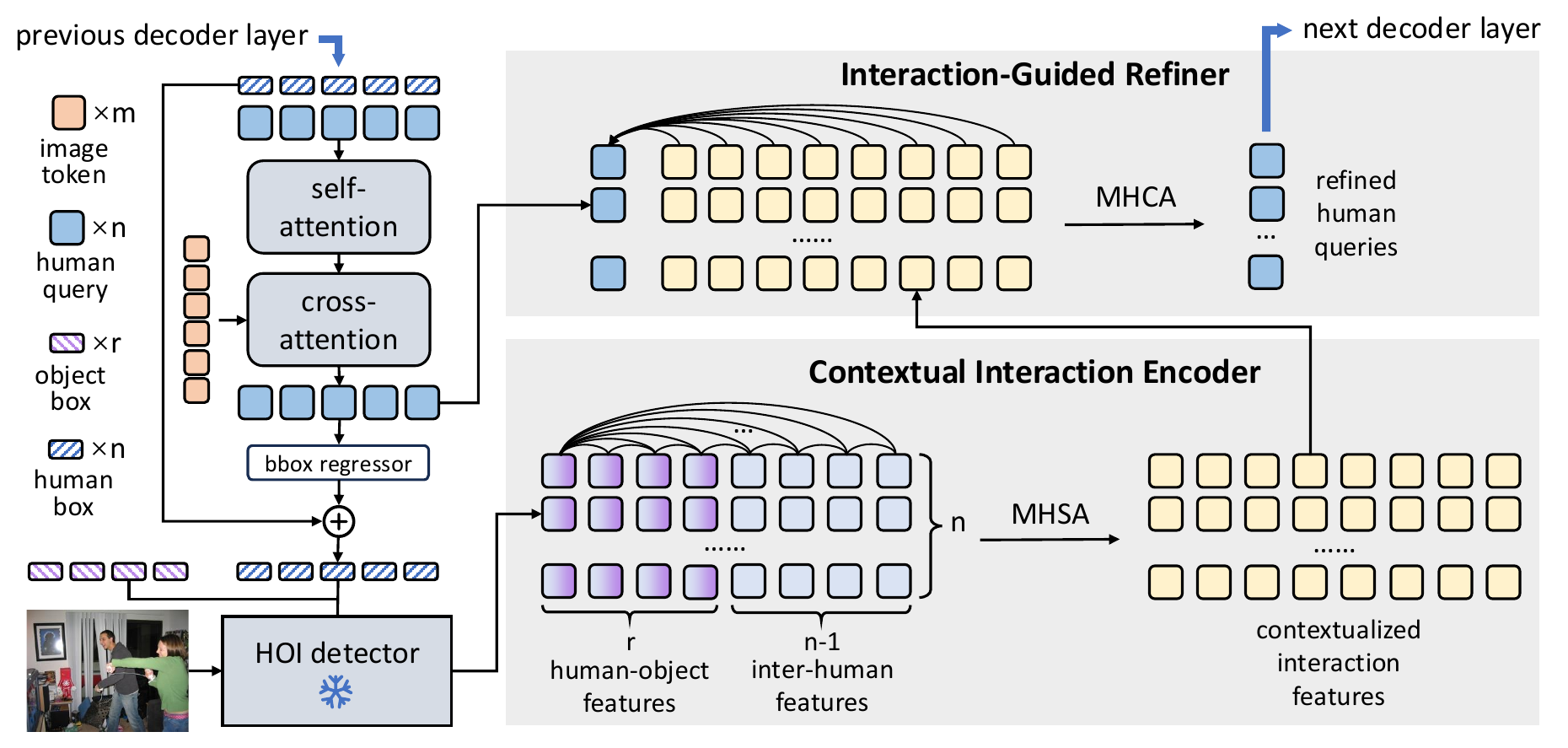}
	\caption{Architecture of a single decoder layer of InterMesh. Each layer takes as input human queries updated by self-attention and cross-attention with image tokens. Human bounding boxes are regressed from these queries and paired with pre-extracted object boxes to form dense human-environment pairs. A pretrained HOI detector extracts interaction features for all pairs. The Contextual Interaction Encoder processes these features using multi-head self-attention (MHSA) to model interdependencies among interactions. The Interaction-Guided Refiner then applies multi-head cross-attention (MHCA) to refine each human query using its associated interaction features. This design enables progressive query refinement guided by interaction context.}
	\label{fig:model}
\end{figure*}

\subsection{Human Model}
We use SMPL \cite{loper2015smpl} as the parametric human body model in this paper, denoted as $\mathcal{M}$. It takes as input the pose parameters $\theta \in \mathbb{R}^{24 \times 3}$, representing the axis-angle rotations of 24 body joints, and the shape parameters $\beta \in \mathbb{R}^{10}$, which control the body proportions in a low-dimensional PCA space. The output is a triangulated mesh of 6890 vertices in 3D space: $V = \mathcal{M}(\theta, \beta) \in \mathbb{R}^{6890 \times 3}$. Then the 3D joints locations can be computed via a predefined linear regressor: $J=\mathcal{J}V$, where $\mathcal{J}\in\mathbb{R}^{K\times6890}$ is the joint regressor and $K$ is the number of joints. 

\subsection{DETR-style End-to-end HMR}
The DETR-style end-to-end human mesh recovery framework consists of an encoder, a decoder, and multiple regression heads for predicting SMPL and camera parameters. Given an input RGB image $I\in\mathbb{R}^{H\times W\times3}$, it is first divided into non-overlapping patches of size $P\times P$. These patches undergo patch embedding and are augmented with positional encoding before being passed through a Transformer encoder, producing a sequence of image feature tokens $T_{img}=\{t_1,t_2,...,t_m\}$, where $m=\frac{H}{P}\times\frac{W}{P}$. In the decoder, a set of learnable human queries $Q_{hum}=\{q_1,q_2,...,q_n\}$ are initialized, each intended to represent a distinct person in the scene. These queries are refined through a stack of Transformer layers that alternate between self-attention and cross-attention. The self-attention layers facilitate message passing among queries, enabling the modeling of inter-person relationships, while the cross-attention layers allow each query to selectively aggregate relevant visual cues from the image tokens $T_{img}$. To improve spatial grounding and focus, the decoder architecture often incorporates the reference points mechanism from Deformable DETR \cite{zhu2020deformable} or DAB-DETR \cite{liu2022dab}. 
% These reference points act as anchors in the image space, guiding attention to relevant regions and improving localization accuracy in dense or cluttered scenes. 
After decoding, each human query is processed independently by dedicated regression heads to estimate the SMPL pose and shape parameters, along with weak-perspective camera parameters $\Pi=(s,t_x,t_y)\in\mathbb{R}^3$, where $s$ is the scale and $t_x,t_y$ are the 2D translations. The final output is a set of 3D human meshes, each projected to the image plane according to its corresponding camera parameters. 

\subsection{HOI Detector}
HOI detection models \cite{zhang2022efficient,lei2023efficient,mao2023clip4hoi,lei2024ez} typically detect all humans and objects in an image, and then classify the interaction type for each human-object pair, often aligning these interactions with natural language descriptions. Importantly, the term \textit{object} in HOI detection broadly refers to both other people and inanimate items, allowing HOI features to encode both inter-human and human-object interactions. To avoid ambiguity, we refer to \textit{objects} in this paper strictly as non-human entities, and use the term \textit{environment} to denote all potential interactive elements in the scene including both people and physical objects. In our implementation, we use EZ-HOI \cite{lei2024ez}, a SOTA zero-shot HOI detector, to extract human-environment interaction features in this paper. 
%  It comprises a visual encoder and a text encoder. The visual encoder processes human-object pairs detected in the image and extracts interaction-centric features with a pretrained CLIP \cite{radford2021learning} visual encoder. Meanwhile, the text encoder transforms detailed HOI class descriptions into embeddings using a CLIP text encoder to align with visual features. Moreover, it leverages relationships between seen and unseen classes for effective zero-shot interaction classification. Since our work focuses on feature-level interaction representations instead of textual outputs, we utilize only the visual encoder. 

\section{InterMesh}
\label{sec:intermesh}

In this section, we detail the architecture of InterMesh. Following prior DETR-style end-to-end HMR models, InterMesh is composed of an $L$-layer encoder, an $L$-layer decoder, and a set of regression heads. The encoder is built upon the design of SAT-HMR \cite{su2024sathmr}, leveraging multi-resolution tokens guided by a scale map to effectively capture visual information at multiple spatial scales. Our key contributions lie in the decoder design and the architecture of a single decoder layer of InterMesh is illustrated in Fig.~\ref{fig:model}. We adopt the reference point mechanism from DAB-DETR \cite{liu2022dab}, where each human query is associated with a bounding box that is iteratively updated across decoder layers. The bounding boxes at each decoder stage are also used as inputs to the visual encoder of the pretrained HOI detector to extract human-environment interaction features. The features are then contextualized and integrated into the decoding process through our proposed Contextual Interaction Encoder and Interaction-Guided Refiner, enabling interaction-aware query refinement throughout the network.

Denote the input human queries of the $l^{th}$ decoder layer as $Q_{hum}^l=\{q_1^l,q_2^l,...,q_n^l\}$ and the corresponding reference points as $p^l=\{p_1^l,p_2^l,...,p_n^l\}$, where $l=0,1,...,L-1$. After self-attention and cross-attention with the image tokens, the queries are updated to their intermediate representations $\tilde{Q}_{hum}^l=\{\tilde{q}_1^l,\tilde{q}_2^l,...,\tilde{q}_n^l\}$. The bounding box for each query is then predicted using a regression head $\mathcal{H}_{bbox}$ as follows:
\begin{equation}
	\begin{aligned} 
        &\mathrm{bh}_i^l = (x_i^l, y_i^l, w_i^l, h_i^l) \\
        &= \mathrm{sigmoid}\left( \mathrm{sigmoid}^{-1}(p_i^l) + \mathcal{H}_{bbox}(\tilde{q}_i^l) \right), i = 1, 2, ..., n 
	\end{aligned}
\end{equation}

The predicted human bounding boxes $\mathrm{Bh}^l=\{\mathrm{bh}_1^l,\mathrm{bh}_2^l,...,\mathrm{bh}_n^l\}$ are then combined with object boxes $\mathrm{Bo}=\{bo_1,bo_2,...,bo_r\}$, which are pre-extracted using an off-the-shelf object detector \cite{carion2020end} from the image. For each human box $\mathrm{bh}_i^l$, we pair it with all other boxes in $\mathrm{Be}^l_i=\{\mathrm{be}_j^l\}_i=\mathrm{Bh}^l\cup \mathrm{Bo}\backslash\{\mathrm{bh}_i^l\}$, yielding a total of $\mathrm{n\cdot(n+r-1)}$ human-object pairs. Each pair $(\mathrm{bh}_i^l,\mathrm{be}_j^l)$ is fed into the HOI detector $\mathcal{D}$ to extract the corresponding interaction feature from the image, which is then projected to the model dimension $d$ via a linear layer $\mathcal{P}$:
\begin{equation}
	t_{i,j}^{l} = \mathcal{P}(\mathcal{D}(I, \mathrm{bh}_i^l, \mathrm{be}_j^l)), i = 1, ..., n, j = 1, ..., n + r - 1
\end{equation}
These features capture both inter-human and human-object interactions and are passed to the Contextual Interaction Encoder and Interaction-Guided Refiner to guide the refinement of human queries.

\subsection{Contextual Interaction Encoder}
Given an intermediate human query $\tilde{q}_i^l$ and its associated interaction features $\{t_{i,j}^l\}_{j=1}^{n+r-1}$, the Contextual Interaction Encoder aims to model the relationships among these interaction features to enrich their contextual understanding. This is achieved by applying a Multi-Head Self-Attention (MHSA) mechanism across all $\mathrm{n+r-1}$ interaction features corresponding to the same human query. The intuition is that both inter-human and human-object interactions surrounding the same individual often influence each other and should be jointly considered. Formally, the enhanced interaction features are computed as:
\begin{equation}
	\tilde{T}_i^l=\mathrm{MHSA}([t_{i,1}^l,t_{i,2}^l,...,t_{i,n+r-1}^l])
\end{equation}
where $\tilde{T}_i^l=\{\tilde{t}_{i,j}^l\}_{j=1}^{n+r-1}\in\mathbb{R}^{(n+r-1)\times d}$ and $d$ is the feature dimension. These refined features capture context-aware interaction semantics for each human query. 

\subsection{Interaction-Guided Refiner}
To further refine the human query representation, the Interaction-Guided Refiner performs Multi-Head Cross-Attention (MHCA) between the original human query $\tilde{q}_i^l$ and its contextually enhanced interaction features $\tilde{T}_i^l$. This mechanism allows each human query to selectively attend to its most informative interactions, thereby incorporating both inter-human and human-object cues into its representation. The refined human query $q_i^{l+1}$ is updated as:
\begin{equation}
	q_i^{l+1}=\mathrm{MHCA}(\tilde{q}_i^l,\tilde{T}_i^l)=\mathrm{MHCA}(\tilde{q}_i^l,\{\tilde{t}_{i,j}^l\}_{j=1}^{n+r-1})
\end{equation}
where the refined query $q_i^{l+1}$ is used as the human query input to the subsequent decoder layer. In the final decoder layer, $q_i^{L}$ is further fed into the prediction heads for human parameter regression. This interaction-guided refinement effectively integrates rich interaction context into the human query, enabling more accurate pose and mesh estimation under complex human-environment settings. 

\subsection{Parameter Regression}
Given the refined human queries of the final decoder layer $Q_{hum}^L$, a set of regression heads $\mathcal{H}=\{\mathcal{H}_{conf}, \mathcal{H}_{shape}, \mathcal{H}_{pose}, \mathcal{H}_{cam}\}$ are applied to predict the confidence score, shape and pose parameters of human model, as well as the camera parameters for all human queries. 
\begin{equation}
    \begin{aligned}
        &y_{i,conf}=\mathcal{H}_{conf}(q_i^L)\\
        &y_{i,shape}=\mathcal{H}_{shape}(q_i^L)\\
        &y_{i,pose}=\mathcal{H}_{pose}(q_i^L)\\
        &y_{i,cam}=\mathcal{H}_{cam}(q_i^L), i=1,2,...,n\\
    \end{aligned}
\end{equation}
Here, the camera parameters follow a weak-perspective formulation, parameterized by a scale factor and 2D translation. Since the predicted scale is inversely proportional to the camera-space depth, we convert the weak-perspective camera parameters to a 3D translation and use the resulting root-joint depth as $y_{i,\mathrm{depth}}$.
The predicted 3D joints and mesh vertices are then projected onto the image plane using the estimated camera parameters to obtain 2D keypoints $y_{i,kpts}$ and bounding boxes $y_{i,bbox}$, which are subsequently used in matching cost and training losses.

\subsection{Matching Cost}
Following DETR, we employ the Hungarian algorithm \cite{kuhn1955hungarian} to perform optimal bipartite matching between predicted outputs and ground-truth targets. The matching cost is a weighted sum of four components: the confidence cost $\mathcal{C}_{conf}$, the bounding box cost $\mathcal{C}_{bbox}$, the Generalized IoU cost $\mathcal{C}_{giou}$ \cite{rezatofighi2019giou}, and the keypoint cost $\mathcal{C}_{kpts}$. Let $y$ denote a prediction and $\hat{y}$ the corresponding ground truth. The individual and the total cost terms are defined as:
\begin{equation}
	\begin{aligned}
		&\mathcal{C}_{conf}=-(1-y_{conf})^{\gamma_{conf}}\cdot\mathrm{log}(y_{conf}) \\
		&\mathcal{C}_{bbox}=\| y_{bbox} - \hat{y}_{bbox} \|_1\\
		&\mathcal{C}_{giou}=- \text{GIoU}(y_{bbox}, \hat{y}_{bbox})\\
		&\mathcal{C}_{kpts}=\| y_{kpts} - \hat{y}_{kpts} \|_1\\
		&\mathcal{C}_{total}=\alpha_{conf}\mathcal{C}_{conf}+\alpha_{bbox}\mathcal{C}_{bbox}+\alpha_{giou}\mathcal{C}_{giou}+\alpha_{kpts}\mathcal{C}_{kpts}
	\end{aligned}
\end{equation}

\subsection{Training Losses}
Following \cite{su2024sathmr}, we employ a combination of loss terms including the scale map loss $\mathcal{L}_{map}$, depth loss $\mathcal{L}_{depth}$, pose parameter loss $\mathcal{L}_{pose}$, shape parameter loss $\mathcal{L}_{shape}$, 3D joint loss $\mathcal{L}_{j3ds}$, 2D joint loss $\mathcal{L}_{j2ds}$, bounding box loss $\mathcal{L}_{box}$, and detection loss $\mathcal{L}_{det}$. Similar to the matching cost, the bounding box loss is composed of an L1 loss and a Generalized IoU (GIoU) loss. The detection loss is implemented as a focal loss, while the remaining terms are standard L1 losses. The overall loss function is defined as a weighted sum of all components:
\begin{equation}
\begin{aligned}
	\mathcal{L}=&\lambda_{map}\mathcal{L}_{map}+\lambda_{depth}\mathcal{L}_{depth}+\lambda_{pose}\mathcal{L}_{pose}+\\
	&\lambda_{shape}\mathcal{L}_{shape}+\lambda_{j3ds}\mathcal{L}_{j3ds}+\lambda_{j2ds}\mathcal{L}_{j2ds}+\\
	&\lambda_{box}\mathcal{L}_{box}+\lambda_{det}\mathcal{L}_{det}
\end{aligned}
\end{equation}

\subsection{Training Parallelism}

\begin{figure}[t!]
	\centering
	\includegraphics[width=\linewidth]{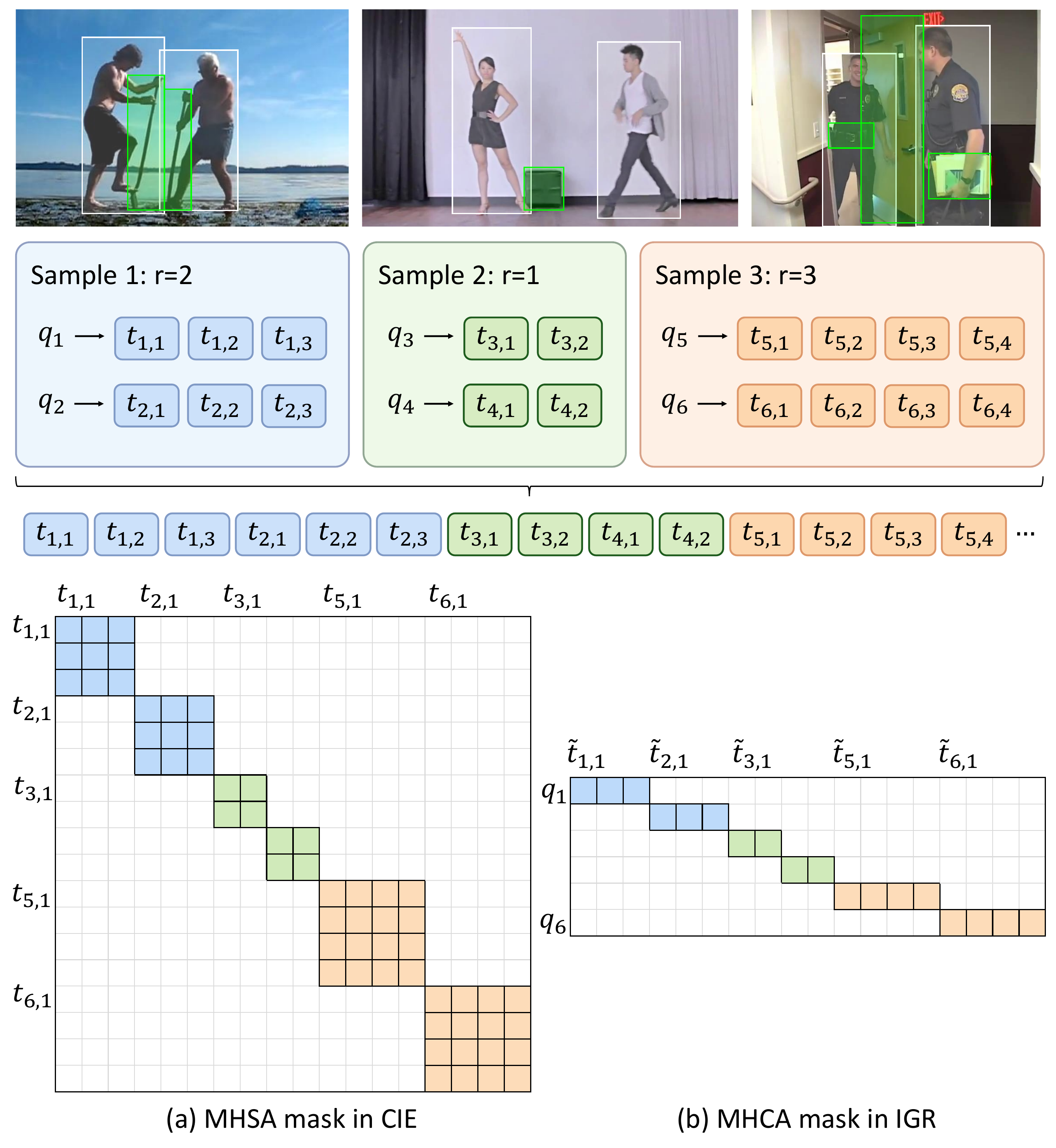}
	\caption{Illustration of the attention masks used for batched training with variable numbers of detected objects, assuming the number of human queries is 2 and the batch size is 3. For sample $b$, each human query is associated with $r_b + n - 1$ interaction tokens, where $r_b$ denotes the number of detected objects. All interaction tokens are then flattened into a single sequence. (a) The MHSA mask in the Contextual Interaction Encoder forms a block-diagonal attention pattern, where each block corresponds to the interaction-token group associated with one human query. (b) The MHCA mask in the Interaction-Guided Refiner constrains each human query to attend only to its corresponding interaction-token segment. Colored regions indicate valid attention, while white regions denote masked-out positions.}
	\label{fig:parallel}
\end{figure}

In practice, each image sample in a training batch may contain a different number of detected object boxes, resulting in a variable number of interaction features for each human query. To enable efficient and fully batched training under such heterogeneity, we design a unified attention masking strategy that allows interaction modeling to proceed in parallel across all samples. Specifically, we first flatten all interaction features for all human queries in the batch into a single long sequence. Then, a block-diagonal attention mask is constructed, where each diagonal block corresponds to the valid attention subspace of a specific human query. 
Fig.~\ref{fig:parallel} illustrates an example of the attention masks when the number of human queries is 2 and the batch size is 3. The three samples contain $r_b=2,1,3$ detected objects, respectively. Each human query is paired with all other human and object boxes (excluding itself), resulting in $r_b + n - 1$ interaction features per query. Consequently, the number of interaction features becomes $3$, $2$, and $4$ for the samples shown. This mechanism ensures that both the MHSA in the Contextual Interaction Encoder and the MHCA in the Interaction-Guided Refiner are restricted to operate only within their respective local interaction contexts, without information leakage across queries or samples.

\section{Experiments}
\label{sec:experiments}

\subsection{Implementation}
We adopt ViT-Base \cite{dosovitskiy2020vit,oquab2023dinov2} as the backbone and set both the encoder and decoder to 6 layers. We use the DETR model pretrained on the HICO-DET \cite{chao2018hicodet} dataset for object detection, with the number of queries set to 100. In our model, the number of human queries is set to $n = 50$. To extract human-environment interaction features, we employ the EZ-HOI model with a ViT-Large backbone, producing feature vectors of dimension 768. These interaction features are then projected into a feature space of dimension $d_{model} = 768$ within our model. The number of heads in the MHSA and MHCA modules is set to 8, and the feed-forward network has a hidden dimension of 2048.
The input image is resized such that its longer edge is 1288 pixels. For data augmentation, we apply random rotations in the range of $[-15^\circ, 15^\circ]$, random scaling by a factor between 0.8 and 2.0, random horizontal flipping with a probability of 0.5, and random cropping with a probability of 0.8.
The weights for the Hungarian matching cost components $\alpha_{conf}$, $\alpha_{bbox}$, $\alpha_{giou}$, and $\alpha_{kpts}$ are set to 0.25, 1.0, 1.0, and 20.0, respectively, with the focal loss parameter $\gamma_{conf}$ set to 2.0. The weighting coefficients for each term in the loss function, $\lambda_{map}$, $\lambda_{depth}$, $\lambda_{pose}$, $\lambda_{shape}$, $\lambda_{j3ds}$, $\lambda_{j2ds}$, $\lambda_{box}$, and $\lambda_{det}$, are set to 4.0, 0.5, 5.0, 3.0, 8.0, 40.0, 2.0, and 1.0, respectively.

To ensure a fair comparison, we follow SAT-HMR \cite{su2024sathmr} for the training datasets and multi-stage strategy. In the first training stage, the model is initialized with the SAT-HMR checkpoint and trained for 3 epochs on a combination of AGORA \cite{patel2021agora}, BEDLAM \cite{black2023bedlam}, COCO \cite{lin2014coco}, MPII \cite{andriluka2014mpii}, CrowdPose \cite{li2019crowdpose}, and Human3.6M \cite{ionescu2013h36m} datasets. In the second stage, we further finetune the model on benchmark datasets such as 3DPW and Hi4D, which are also used for final evaluation. For both stages, we use AdamW optimizer \cite{loshchilov2017adamw} with a learning rate of 1e-5 and weight decay of 1e-4. Training is conducted on two NVIDIA A800 GPUs with a total batch size of 40. Data augmentation techniques include random rotation, horizontal flipping, scaling and cropping. During evaluation, we filter out low-confidence detections using a threshold of 0.3. Evaluation metrics for the 3DPW, CMU Panoptic, Hi4D and CHI3D include Mean Per-Joint Position Error (MPJPE), Procrustes Aligned MPJPE (PA-MPJPE), and Per-Vertex Error (PVE), all reported in millimeters (mm). For the MuPoTS dataset, we report 3D Percentage of Correct Keypoints (PCK). 

\subsection{Quantitative Comparison}

\begin{table*}[htbp]
	\centering
    \caption{Comparison with SOTA methods on 3DPW, MuPoTS, and CMU Panoptic datasets. \textbf{Bold} indicates the best result, and \underline{underlined} indicates the second-best. $\downarrow$: lower is better; $\uparrow$: higher is better.}
	\begin{tabular}{l|ccc|cc|c}
		\toprule
		\multicolumn{1}{l}{\multirow{2}{*}{\textbf{Method}}} & \multicolumn{3}{|c|}{\textbf{3DPW}} & \multicolumn{2}{c|}{\textbf{MuPoTS}} & \multicolumn{1}{c}{\textbf{CMU Panoptic}} \\
		\multicolumn{1}{c}{} & \multicolumn{1}{|c}{PA-MPJPE $\downarrow$} & \multicolumn{1}{c}{MPJPE $\downarrow$} & \multicolumn{1}{c|}{PVE $\downarrow$} & \multicolumn{1}{c}{PCK (All) $\uparrow$} & \multicolumn{1}{c|}{PCK (Match) $\uparrow$} & \multicolumn{1}{c}{MPJPE $\downarrow$} \\  \midrule
		CRMH \cite{jiang2020coherent} & - & - & - & 69.1 & 72.2 & 143.2 \\
		3DCrowdNet \cite{choi2022learning} & 51.5 & 81.7 & 98.3 & 72.7 & 73.3 & 127.6 \\
		ROMP \cite{sun2021monocular} & 47.3 & 76.6 & 93.4 & 69.9 & 72.2 & 128.2 \\
		BEV \cite{sun2022putting} & 46.9 & 78.5 & 92.3 & 70.2 & 75.2 & 109.5 \\
		PSVT \cite{qiu2023psvt} & 45.7 & 75.5 & 84.9 & - & - & 105.7 \\
		Multi-HMR \cite{baradel2024multi} & 41.7 & \underline{61.4} & 75.9 & 85.0 & 89.3 & 96.5 \\
		SAT-HMR \cite{su2024sathmr} & \underline{41.6} & 63.6 & \underline{73.7} & \underline{89.0} & \underline{90.1} & \underline{84.2} \\
		\rowcolor{gray!20}
		Ours & \textbf{39.8} & \textbf{60.7} & \textbf{71.0} & \textbf{89.9} & \textbf{90.3} & \textbf{75.9} \\ \bottomrule
		% SAT-HMR* \cite{su2024sathmr} & \underline{38.2} & \underline{61.3} & \underline{71.1} & \underline{87.4} & \underline{88.2} & \underline{80.1} \\
		% \rowcolor{gray!20} Ours* & \textbf{37.4} & \textbf{60.3} & \textbf{69.7} & \textbf{90.1} & \textbf{90.5} & \textbf{75.7} \\ \bottomrule
	\end{tabular}
	\label{tab:3dpw_mupots_panoptic_results}
\end{table*}

\begin{table}[htbp]
	\centering
    \caption{Detailed comparison with SOTA methods on four scenes of CMU Panoptic in terms of MPJPE.}
	\begin{tabular}{l|cccc}
		\toprule
		\multicolumn{1}{l}{\multirow{2}{*}{\textbf{Method}}} & \multicolumn{4}{|c}{\textbf{CMU Panoptic}} \\
		\multicolumn{1}{c}{} & \multicolumn{1}{|c}{Haggling} & Mafia & Ultimatum & Pizza \\ \midrule
		CRMH \cite{jiang2020coherent} & 129.6 & 133.5 & 153.0 & 156.7 \\
		3DCrowdNet \cite{choi2022learning} & 109.6 & 135.9 & 129.8 & 135.6 \\
		ROMP \cite{sun2021monocular} & 110.8 & 122.8 & 141.6 & 137.6 \\
		BEV \cite{sun2022putting} & 90.7 & 103.7 & 113.1 & 125.2 \\
		PSVT \cite{qiu2023psvt} & 88.7 & 97.9 & 115.2 & 121.2 \\
		SAT-HMR \cite{su2024sathmr} & \underline{67.9} & \underline{78.5} & \underline{95.8} & \underline{94.6} \\
		\rowcolor{gray!20}
		Ours & \textbf{60.2} & \textbf{73.5} & \textbf{84.9} & \textbf{83.3} \\ \bottomrule
		% SAT-HMR* \cite{su2024sathmr} & \underline{62.2} & \underline{78.4} & \underline{89.6} & \underline{87.7} \\
		% \rowcolor{gray!20} Ours* & \textbf{60.2} & \textbf{74.0} & \textbf{85.2} & \textbf{82.0} \\ \bottomrule
	\end{tabular}
	\label{tab:detailed_panoptic_results}
\end{table}

Table~\ref{tab:3dpw_mupots_panoptic_results} presents the experimental results on 3DPW, MuPoTS, and CMU Panoptic datasets. On the 3DPW dataset, our method achieves a 100\% recall, ensuring that all detected instances are included in the evaluation of the three mesh accuracy metrics. Compared to the previous best method, InterMesh reduces PA-MPJPE by 1.8mm (4.3\%), MPJPE by 0.7mm (1.1\%), and PVE by 2.7mm (3.7\%), demonstrating a consistent and significant improvement across all metrics. On the MuPoTS dataset, although prior methods have already achieved high accuracy, our approach still brings further gains, improving PCK (match) by 0.2\% and PCK (all) by 0.9\% respectively. On the CMU Panoptic dataset, InterMesh reduces the average MPJPE by 8.3mm (9.9\%), marking a substantial improvement. 
Table~\ref{tab:detailed_panoptic_results} shows the detailed comparison on four scenes of CMU Panoptic. Notably, in the challenging and crowded "pizza" scene, it achieves a reduction of 11.3mm in MPJPE, highlighting the effectiveness of InterMesh in handling complex human-object and human-human interaction scenarios.

\begin{table}[t!]
	\centering
    \caption{Comparison with SOTA methods on Hi4D.}
	\begin{tabular}{l|ccc}
		\toprule
		\multirow{2}{*}{\textbf{Method}} & \multicolumn{3}{c}{\textbf{Hi4D}} \\
		& MPJPE $\downarrow$ & PA-MPJPE $\downarrow$ & PVE $\downarrow$ \\ \midrule
		CLIFF \cite{li2022cliff} & 91.3 & 53.6 & 109.6 \\
		4D-Humans \cite{goel2023humans} & 72.1 & 52.4 & 88.6 \\
		BEV \cite{sun2022putting} & 91.8 & 52.2 & 101.2 \\
		TRACE \cite{sun2023trace} & 83.8 & 60.4 & - \\
		GroupRec \cite{huang2023reconstructing} & 82.4 & 51.6 & 88.6 \\
		BUDDI \cite{muller2024generative} & 96.8 & 70.6 & 116.0 \\
		CloseInt \cite{huang2024closely} & 63.1 & 47.5 & 76.4 \\
		Fang \textit{et al.} \cite{fang2024capturing} & 75.0 & 59.7 & - \\
		SAT-HMR \cite{su2024sathmr} & \underline{51.1} & \underline{38.2} & \underline{62.5} \\
		ReconClose \cite{huang2025reconstructing} & 59.1 & 44.3 & 72.0 \\
		\rowcolor{gray!20}
		Ours & \textbf{46.9} & \textbf{35.6} & \textbf{57.0} \\ \bottomrule
	\end{tabular}
	\label{tab:hi4d_results}
\end{table}

\begin{table}[t!]
	\centering
    \caption{Comparison with SOTA methods on CHI3D.}
	\begin{tabular}{l|cc}
		\toprule
		\multirow{2}{*}{\textbf{Method}} & \multicolumn{2}{c}{\textbf{CHI3D}} \\
		& MPJPE $\downarrow$ & PA-MPJPE $\downarrow$ \\ \midrule
		REMIPS \cite{fieraru2021remips} & 84.1 & 55.9 \\
		BEV \cite{sun2022putting} & 89.1 & 54.6 \\
		CLIFF \cite{li2022cliff} & 72.4 & \textbf{47.5} \\
		4D-Humans \cite{goel2023humans} & 73.1 & 48.6 \\
		TRACE \cite{sun2023trace} & 75.9 & 49.7 \\
		Fang \textit{et al.} \cite{fang2024capturing} & \underline{71.9} & \underline{47.8} \\
		% SAT-HMR \cite{su2024sathmr} &  &  \\
		\rowcolor{gray!20} Ours & \textbf{69.5} & 48.0 \\ \bottomrule
	\end{tabular}
	\label{tab:chi3d_results}
\end{table}

Table~\ref{tab:hi4d_results} and \ref{tab:chi3d_results} show the results on two interacted human datasets: Hi4D and CHI3D. On Hi4D, InterMesh significantly outperforms existing approaches for interactive human reconstruction, achieving reductions in MPJPE, PA-MPJPE, and PVE by 4.2mm (8.2\%), 2.6mm (6.8\%), and 5.5mm (8.8\%), respectively. On CHI3D, InterMesh achieves a 2.4mm reduction of MPJPE and comparable PA-MPJPE to previous SOTA. These results further demonstrate the effectiveness of our method.

\subsection{Visualization Results}

\begin{figure*}[htbp]
	\centering
	\includegraphics[width=\textwidth]{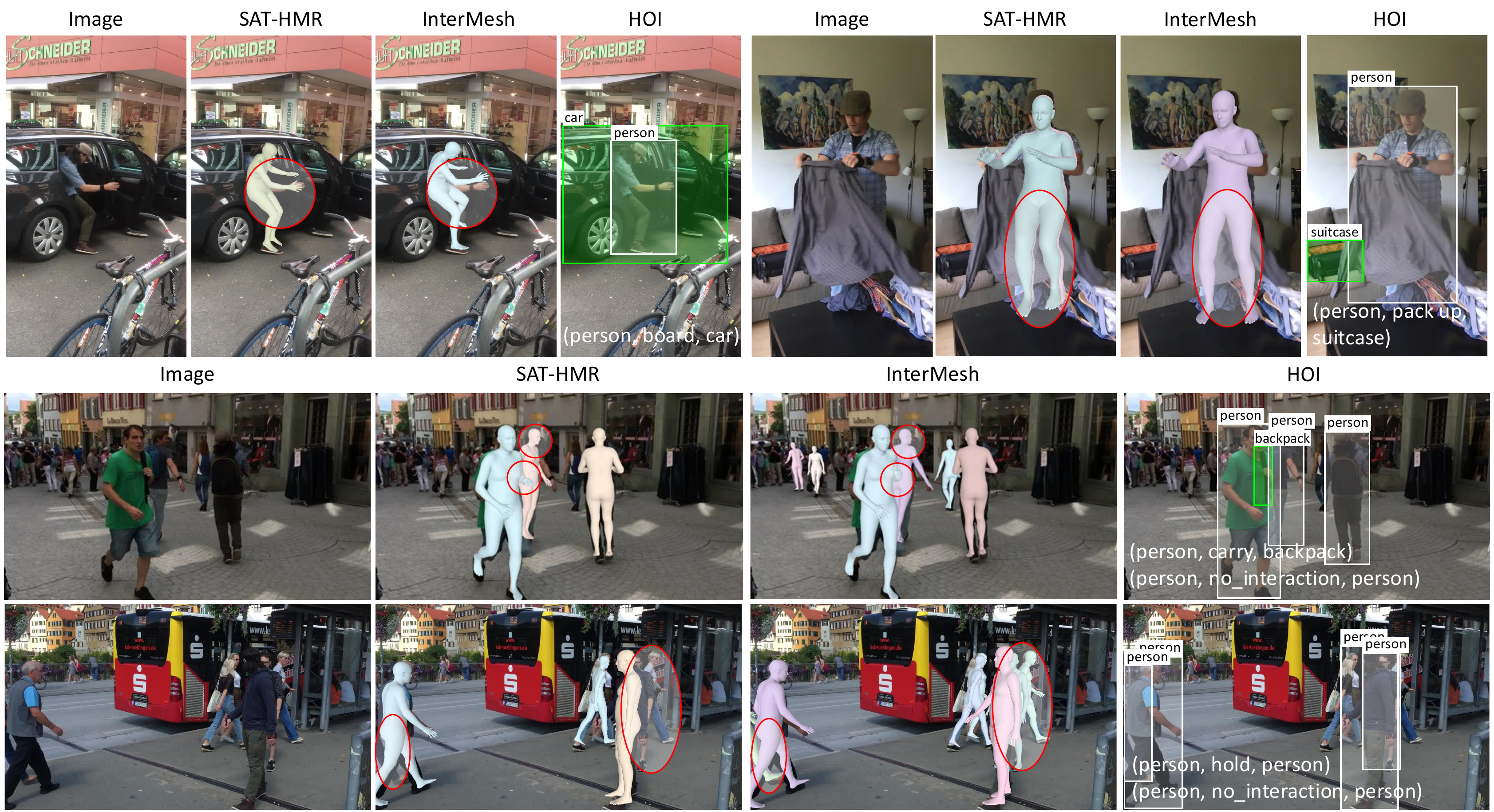}
	\caption{Qualitative comparison on 3DPW.}
	\label{fig:sample_vis_3dpw}
\end{figure*}

\begin{figure}[htbp]
	\centering
	\includegraphics[width=\linewidth]{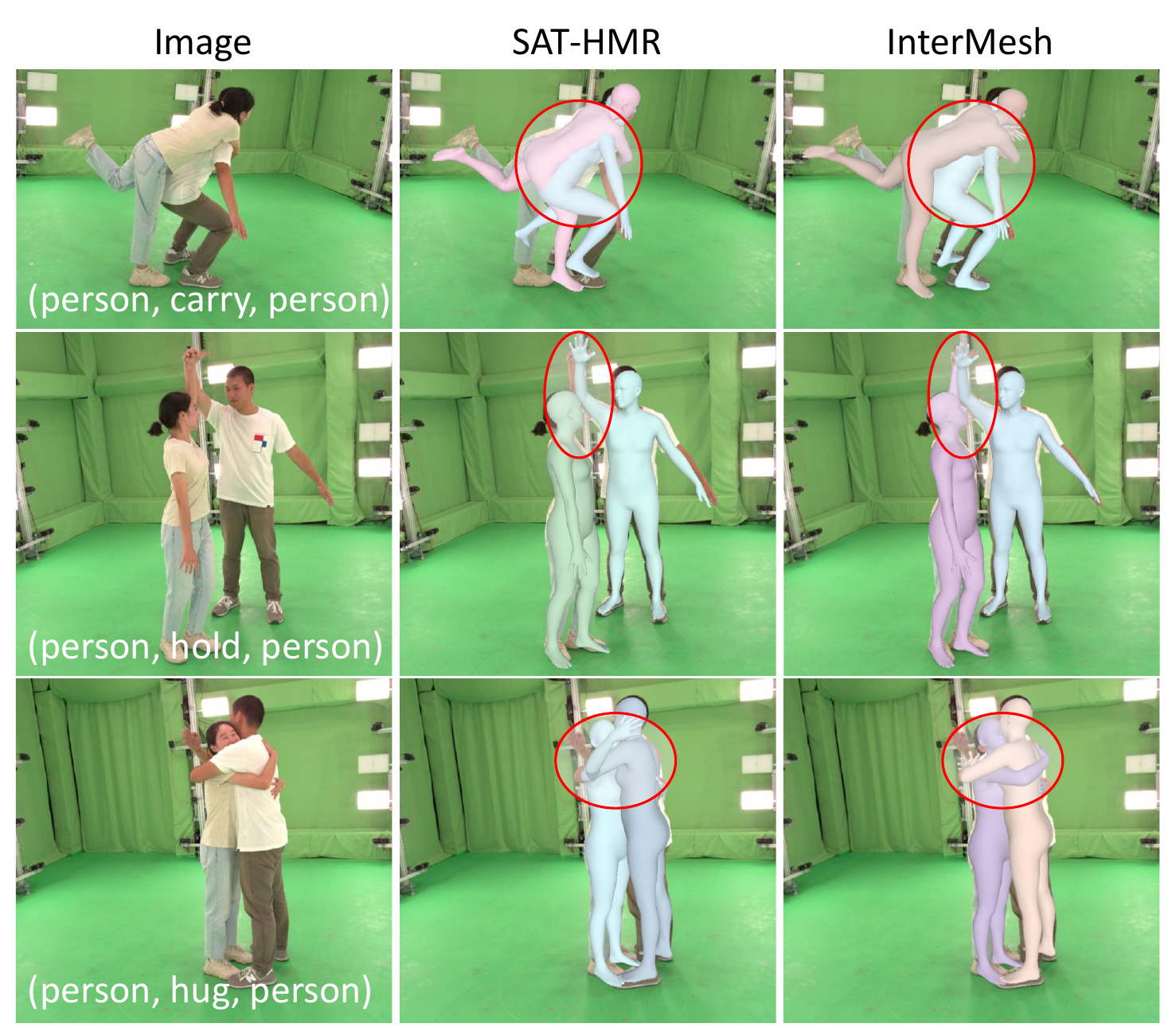}
	\caption{Qualitative comparison on Hi4D.}
	\label{fig:sample_vis_hi4d}
\end{figure}

% \begin{figure*}
% 	\centering
% 	\includegraphics[width=0.9\textwidth]{figures/appendix_vis_3dpw.pdf}
% 	\caption{More visualization results on 3DPW.}
% 	\label{fig:appendix_vis_3dpw}
% \end{figure*}

\begin{figure*}
	\centering
	\includegraphics[width=\textwidth]{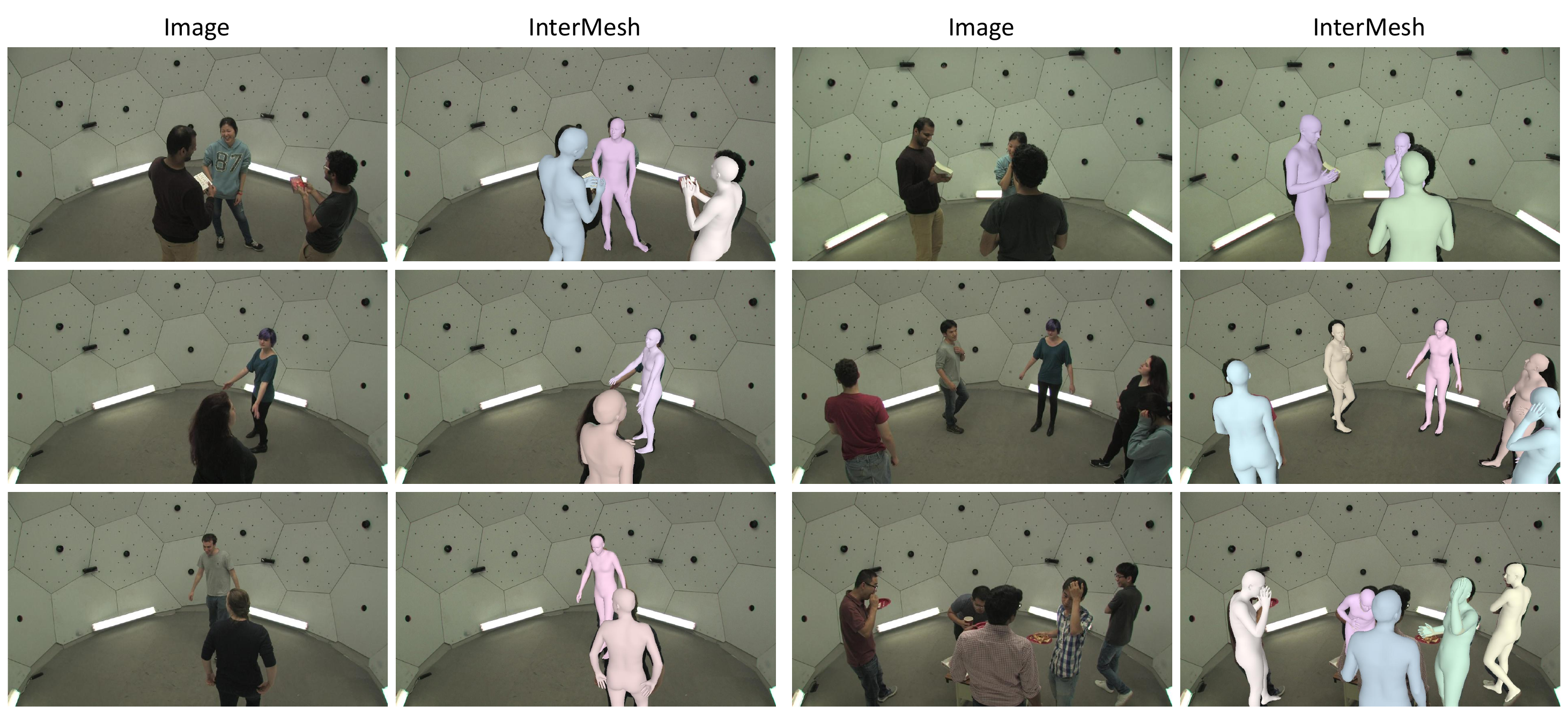}
	\caption{Visualization results on CMU Panoptic.}
	\label{fig:appendix_vis_panoptic}
\end{figure*}

\begin{figure*}
	\centering
	\includegraphics[width=\textwidth]{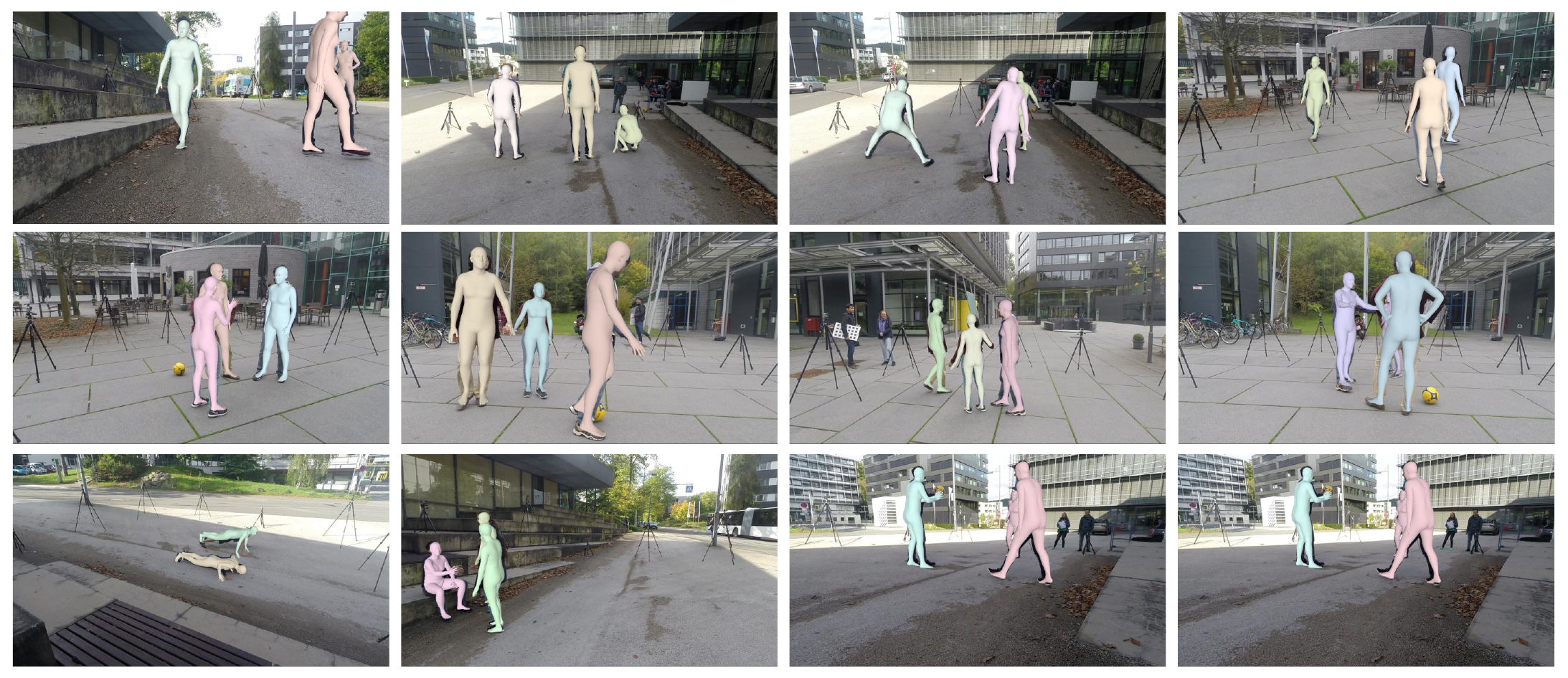}
	\caption{Visualization results on MuPoTS.}
	\label{fig:appendix_vis_mupots}
\end{figure*}

\begin{figure*}
	\centering
	\includegraphics[width=\textwidth]{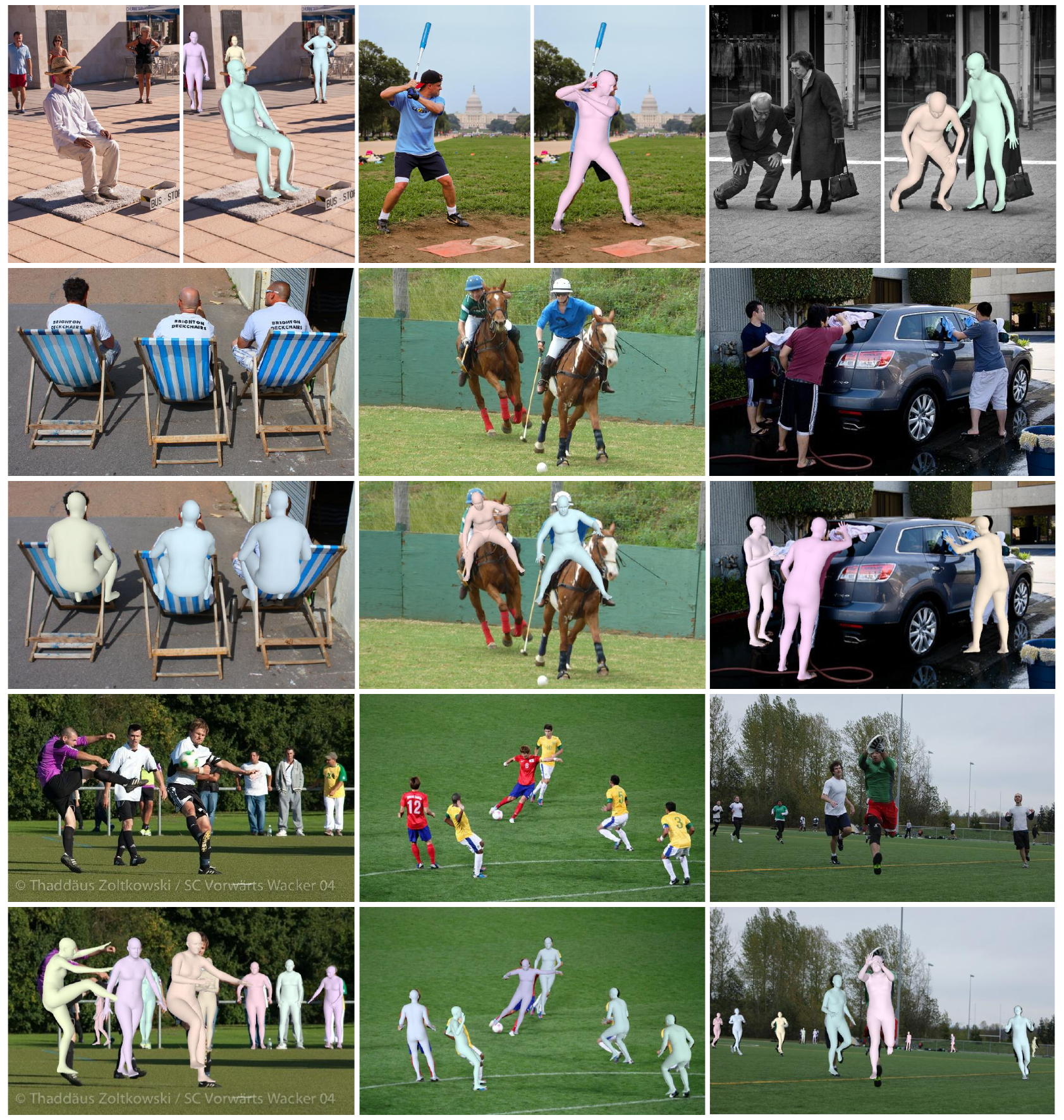}
	\caption{Visualization results on in-the-wild samples.}
	\label{fig:appendix_vis_hicodet}
\end{figure*}

Fig.~\ref{fig:sample_vis_3dpw} shows a qualitative comparison between InterMesh and SAT-HMR on the 3DPW dataset. By leveraging structured interaction features, InterMesh consistently delivers more accurate and context-aware predictions than the baseline. For example, in the first sample where a person is getting out of a car, the HOI detector identifies a “boarding” interaction. This semantic guidance enables InterMesh to produce a mesh that more faithfully reflects the actual human pose and its surrounding context. In the second sample, the suitcase in the scene provides a packing-up activity prior, which favors a more upright standing pose. This helps reduce failure cases where occlusions lead to implausible lower-body configurations such as over-bent knees. In crowded scenes, such as the third and fourth examples, InterMesh not only reconstructs more realistic meshes but also improves the accuracy of human detection. 

Fig.~\ref{fig:sample_vis_hi4d} presents the results of InterMesh on the Hi4D dataset, which contains challenging scenarios with close inter-person interactions and severe occlusions. By explicitly modeling inter-human interaction cues, InterMesh is able to better capture the spatial relationships and behavioral dependencies between nearby individuals. For instance, in interactions such as one person carrying another, hand-holding, and hugging, the interaction features provide additional semantic constraints that help disambiguate body orientations and limb configurations under heavy overlap. As a result, InterMesh produces more accurate and physically plausible multi-person poses and body meshes compared with methods relying solely on implicit query interactions.

Additional qualitative results on benchmarks and in-the-wild samples are shown in Fig.~\ref{fig:appendix_vis_panoptic}, \ref{fig:appendix_vis_mupots}, and \ref{fig:appendix_vis_hicodet}. As illustrated, InterMesh demonstrates high accuracy in scenes with close human-object interactions, while also maintaining robust performance across a diverse set of challenging scenarios.

\begin{figure}[htbp]
    \centering
    \includegraphics[width=\linewidth]{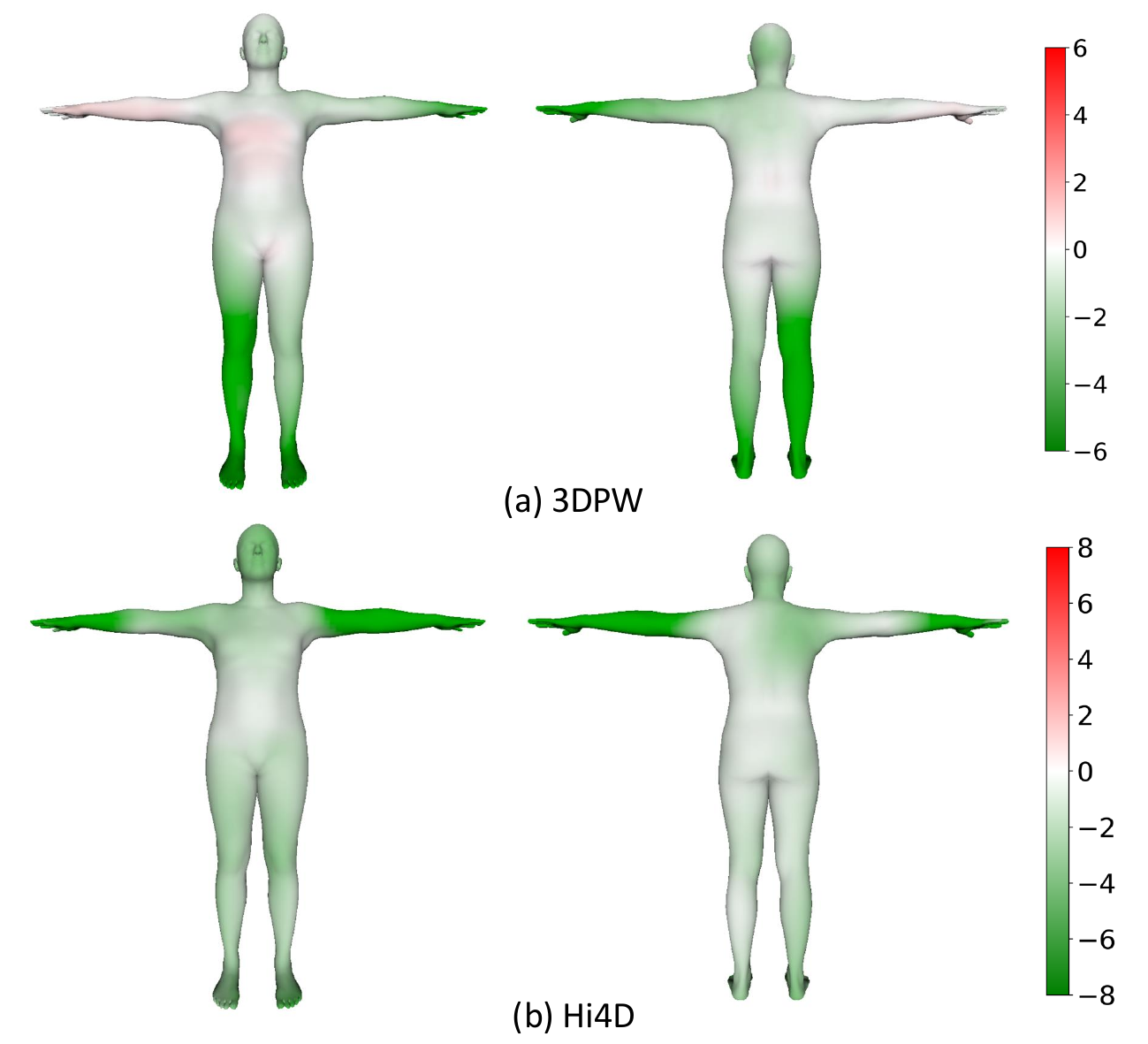}
    \caption{Per-vertex error comparison on 3DPW and Hi4D. Green regions indicate where InterMesh achieves lower errors than the baseline, while red highlights areas with increased errors.}
    \label{fig:vertex_vis}
\end{figure}

Fig.~\ref{fig:vertex_vis} presents a per-vertex error comparison between InterMesh and SAT-HMR on the 3DPW and Hi4D dataset. InterMesh achieves clear improvements on distal body parts such as the head, hands, and feet, which are more frequently involved in human-environment interactions than the torso. This suggests that explicitly modeling interaction context helps the model focus on body regions most affected by external contacts and activities.

\subsection{Ablation Study}

To evaluate the effectiveness of interaction features and the proposed components, Contextual Interaction Encoder (CIE) and Interaction-Guided Refiner (IGR), we conduct an ablation study on the Hi4D dataset, as summarized in Table~\ref{tab:ablation_module}. The first two rows compare SAT-HMR and InterMesh when directly evaluated on Hi4D, showing that explicit interaction modeling enables better generalization to this previously unseen benchmark. We further assess the contribution of each component. In particular, the fourth row shows a variant of our method where human queries directly cross-attend to interaction features from the HOI detector, without contextualization via the CIE. Although this variant outperforms the baseline (third row), it still lags behind the full InterMesh model (bottom row), underscoring the importance of contextualizing raw interaction features.

To separate the contributions of human-human and human-object interaction features, we conduct an ablation study on 3DPW. Specifically, we evaluate four variants: (i) no interaction features; (ii) human-human features only; (iii) human-object features only; (iv) both features. As shown in Table~\ref{tab:ablation_human_vs_object}, explicitly modeling interaction features consistently improves performance over the baseline without interaction modeling. Introducing human-human features alone yields a moderate improvement, indicating that explicit interaction representation is beneficial. Moreover, incorporating human-object features leads to a larger performance gain, highlighting the importance of object context that is entirely absent in prior multi-person HMR methods. Combining both interaction types achieves the best results across all metrics, demonstrating that human-human and human-object cues provide complementary information for more accurate pose and shape estimation.

\begin{table}[t!]
	\centering
    \caption{Ablation study on components of InterMesh on Hi4D.}
	\begin{tabular}{c|ccc|ccc}
		\toprule
		& CIE & IGR & Finetune & MPJPE $\downarrow$ & PA-MPJPE $\downarrow$ & PVE $\downarrow$ \\ 
		\midrule
		1 & \ding{55} & \ding{55} & \ding{55} &  93.4  &  61.8  &  110.1  \\
		2 & \ding{51} & \ding{51} & \ding{55} &  91.1  &  60.9  &  108.3  \\
		3 & \ding{55} & \ding{55} & \ding{51} &  51.1 &  38.2  &  62.5  \\
		4 & \ding{55} & \ding{51} & \ding{51} &  48.6  &  36.4  &  59.3  \\
		5 & \ding{51} & \ding{51} & \ding{51} &  46.9  &  35.6  &  57.0  \\ 
		\bottomrule
	\end{tabular}
	\label{tab:ablation_module}
\end{table}

\begin{table}[t!]
    \centering
    \caption{Ablation study on interaction feature types on 3DPW, comparing human-object features, human-human features, and their combination.}
    \begin{tabular}{l|ccc}
        \toprule
        Feature Type & PA-MPJPE $\downarrow$ & MPJPE $\downarrow$ & PVE $\downarrow$ \\
        \midrule
        None & 41.6 & 63.6 & 73.7 \\
        Human-human only & 41.1 & 63.0 & 73.1 \\
        Human-object only & 40.2 & 61.6 & 71.8 \\
        Both & 39.8 & 60.7 & 71.0 \\
        \bottomrule
    \end{tabular}
    \label{tab:ablation_human_vs_object}
\end{table}

\begin{figure}[t!]
	\centering
	\includegraphics[width=0.8\linewidth]{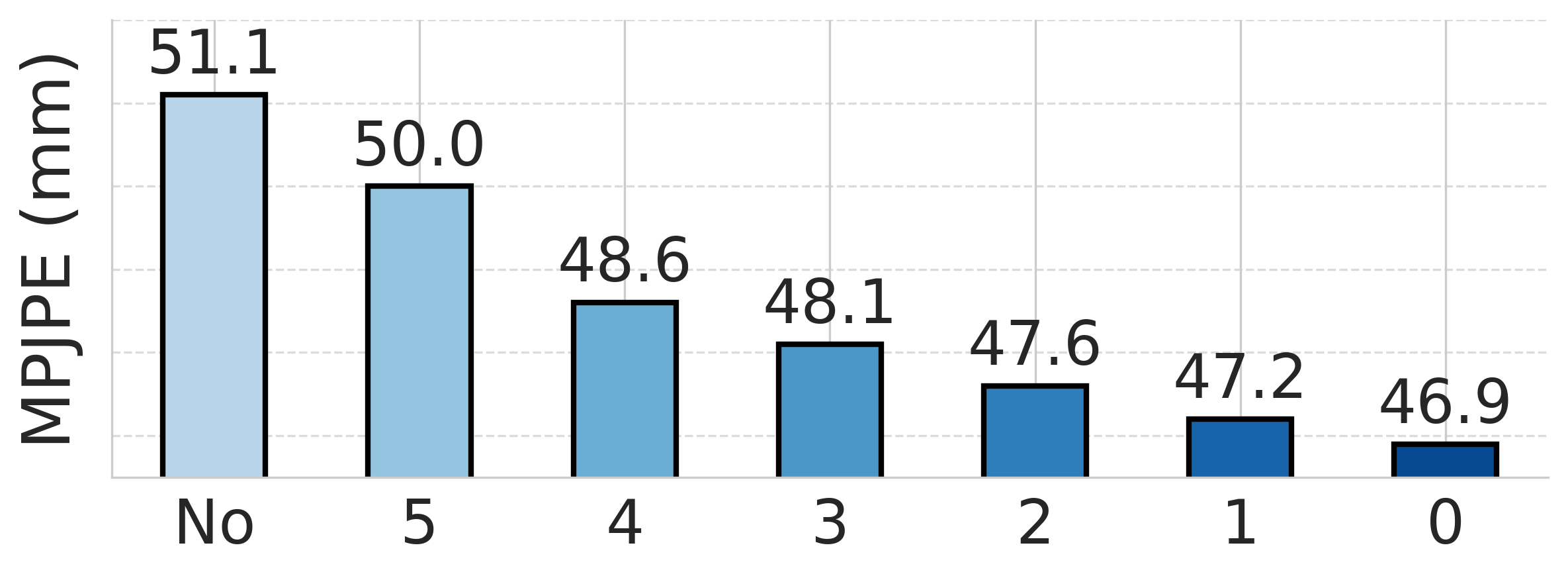}
	\caption{Ablation study on the starting decoder layer for incorporating human-environment interaction features. }
	\label{fig:ablation_least_lvl}
\end{figure}

We further investigate the impact of incorporating human-environment interaction features from different depths of the decoder. As shown in Fig.~\ref{fig:ablation_least_lvl}, performance improves as these modules are introduced from earlier decoder layers. This suggests that deeper integration of interaction cues throughout the decoding process leads to better performance.

\begin{table*}[htbp]
    \centering
    \begin{tabularx}{0.75\linewidth}{l|>{\centering\arraybackslash}X>{\centering\arraybackslash}X|>{\centering\arraybackslash}X>{\centering\arraybackslash}X|>{\centering\arraybackslash}X>{\centering\arraybackslash}X}
        \toprule
        \multirow{2}{*}{\textbf{Method}} &
        \multicolumn{2}{c|}{\textbf{Parameters (M)}} &
        \multicolumn{2}{c|}{\textbf{FLOPs (G)}} &
        \multicolumn{2}{c}{\textbf{Runtime (ms)}} \\
        & HMR Model & HOI Detector & HMR Model & HOI Detector & HMR Model & HOI Detector \\
        \midrule
        SAT‑HMR & 221.9 & — & 449.2 & — & 42.0 & — \\
        InterMesh & 272.7 & 440.3 & 644.4 & 2152.8 & 49.0 & 112.8 \\
        \bottomrule
    \end{tabularx}
    \caption{Comparison of model complexity and computational cost between InterMesh and SAT-HMR. Model parameters, FLOPs and runtime are reported separately for the HMR model and the external HOI detector during inference on one RTX 4090 GPU.}
    \label{tab:appendix_complexity}
\end{table*}

\subsection{Computational Cost}

We evaluate the computational cost of our proposed InterMesh model in comparison with the baseline model SAT-HMR. Table~\ref{tab:appendix_complexity} summarizes the number of model parameters, FLOPs, and runtime for a single image on one RTX 4090 GPU. InterMesh incurs higher computational cost than SAT‑HMR mainly due to the additional HOI detection component. However, compared to methods such as SAT-HMR that only estimate human meshes, InterMesh also produces fine-grained HOI labels, which provides additional interpretable interaction outputs. The increase in complexity within the HMR model itself is moderate. This allows InterMesh to directly benefit from future improvements in lightweight interaction detection without modifying the core mesh recovery architecture.

\section{Limitations}
While InterMesh achieves strong performance by incorporating structured interaction semantics, it currently processes each frame independently and does not explicitly model temporal consistency across video sequences. Incorporating temporal cues could further enhance the stability and realism of mesh predictions, especially in dynamic scenes. Additionally, InterMesh adopts a two-stage approach where human-environment interaction features are extracted separately via an external HOI detector. This design incurs additional computational overhead. A promising future direction is to develop a unified framework that jointly performs HOI detection and mesh recovery in a single stage. Such an approach could reduce inference latency and enable joint optimization of both tasks. Realizing this vision would benefit from large-scale datasets annotated with both SMPL parameters and fine-grained HOI labels, paving the way for more efficient and integrated 3D human understanding.

\section{Conclusion}
We propose InterMesh, a new framework that brings explicit modeling of human-environment interactions into the task of human mesh recovery. By integrating interaction features extracted from a pretrained HOI detector, our method moves beyond purely appearance-based reasoning and introduces a structured understanding of the scene context. We further design two modules, Contextual Interaction Encoder and Interaction-Guided Refiner, to contextualize and selectively absorb interaction semantics into human queries. Through extensive experiments, we demonstrate that InterMesh significantly improves performance across multiple benchmarks, validating the benefit of reasoning about humans as part of an interactive environment. We hope this work inspires future efforts toward more holistic and scene-aware human reconstruction.

\bibliographystyle{IEEEtran}
\bibliography{manuscript}

\end{document}